\pdfoutput=1
\documentclass[11pt]{article}

\usepackage{emnlp2021}

\usepackage{times}
\usepackage{latexsym}

\usepackage[T1]{fontenc}
\usepackage[utf8]{inputenc}

\usepackage{microtype}

\usepackage{amssymb}
\usepackage{amsfonts}
\usepackage{amsmath}

\usepackage{makecell}
\usepackage{multirow}
\usepackage{graphicx}
\usepackage{subcaption}

\usepackage{CJKutf8}

\newcommand*{\affaddr}[1]{#1}
\newcommand*{\affmark}[1][*]{\textsuperscript{#1}}
\newcommand*{\email}[1]{\texttt{#1}}

\title{Unsupervised Cross-lingual Adaptation for Sequence Tagging and Beyond}

\author{
Xin Li\affmark[1,2], Lidong Bing\affmark[2], Wenxuan Zhang\affmark[1], Zheng Li\affmark[3] and
Wai Lam\affmark[1]\\
\affaddr{\affmark[1]The Chinese University of Hong Kong}\\
\affaddr{\affmark[2]DAMO Academy, Alibaba Group}\\
\affaddr{\affmark[3]Hong Kong University of Science and Technology}\\
\email{\{lixin,wxzhang,wlam\}@se.cuhk.edu.hk}\\
\email{l.bing@alibaba-inc.com}\\
\email{zlict@connect.ust.hk}\\
 }

\begin{document}
\maketitle
\begin{abstract}
Cross-lingual adaptation with multilingual pre-trained language models (mPTLMs) mainly consists of two lines of works: zero-shot approach and translation-based approach, which have been studied extensively on the sequence-level tasks. We further verify the efficacy of these cross-lingual adaptation approaches by evaluating their performance on more fine-grained sequence tagging tasks. After re-examining their strengths and drawbacks, we propose a novel ``\textit{\textbf{warmup-then-adaptation}}'' framework to better exploit the translated training sets while inheriting the cross-lingual capability of the mPTLMs. Instead of simply augmenting the source-language training data with the machine-translated data, we tailor-make a warmup mechanism to distill multilingual task-specific knowledge from the translated data in each language and inject them into the model. Then, an adaptation approach is applied to the refined model parameters and the cross-lingual transfer is performed in a warm-start way. The experimental results on nine target languages over three diverse sequence tagging tasks demonstrate that our method is beneficial to the cross-lingual adaptation with mPTLMs.
\end{abstract}

\section{Introduction}
The emergence of multilingual pre-trained language models (\textbf{mPTLMs})\footnote{We abbreviate ``pre-trained language models'' as ``PTLMs'' rather than ``PLMs'' to differentiate it with ``probabilistic language models''~\cite{kneser1995improved,bengio2003neural}.}~\cite{devlin-etal-2019-bert,mulcaire-etal-2019-polyglot,lample2019cross,conneau-etal-2020-unsupervised} has led to significant performance gains on a variety of cross-lingual natural language understanding (\textbf{XLU}) tasks~\cite{lewis2019mlqa,conneau-etal-2018-xnli,hu2020xtreme}. Similar to 
the monolingual scenario~\cite{radford2018improving,peters-etal-2018-deep,yang2019xlnet,liu2019roberta}, exploiting mPTLMs for cross-lingual transfer\footnote{Without specification, ``cross-lingual transfer/adaptation'' in this paper refers to unsupervised transfer where neither labeled data in target language nor parallel corpus is available.} usually involves two phases: 1) pre-train mPTLMs on a large multilingual corpus, and 2) adapt the pre-trained mPTLMs and task-specific layers to the target language. Intuitively, advancing either pre-training technique or adaptation approach is useful for improving the XLU performance. However, due to the prohibitive computational cost of pre-training on a large-scale corpus, designing a better adaptation framework is more practical to the NLP community. 

Among the research efforts on cross-lingual adaptation, the most widely-used approach is zero-shot adaptation~\cite{pires-etal-2019-multilingual,wu-dredze-2019-beto,keung-etal-2019-adversarial}, where the model with a mPTLM backbone is solely fine-tuned on the labeled data from the source language (typically English). Then, with the help of the multilingual pre-training, the fine-tuned model is applied seamlessly on the testing data of the target language. Another line of works belongs to translation-based adaptation~\cite{conneau-etal-2020-unsupervised,artetxe2019massively,yang-etal-2019-paws,eisenschlos-etal-2019-multifit,huang-etal-2019-unicoder,cao2020multilingual}, whose core idea is to borrow machine translation techniques to translate the source-language training set into the target language~\cite{banea-etal-2008-multilingual,duh-etal-2011-machine,tiedemann-etal-2014-treebank}. The cross-lingual adaptation is achieved via supervised training on the translated data. 

Despite the superiority of translation-based adaptation to zero-shot adaptation on text classification~\citep{prettenhofer-stein-2010-cross,schwenk-li-2018-corpus} and text pair classification~\citep{conneau-etal-2018-xnli,liu-etal-2019-xqa,yang-etal-2019-paws}, it still remains unknown which one is better on more fine-grained XLU tasks, such as named entity recognition. One reason is that most of the existing translation-based approaches, built on top of off-the-shelf translators, do not have access to token-level alignment for building pseudo-labeled data in the target language. Therefore, they fail to be applied to more fine-grained XLU tasks. Although some findings from existing works suggest that the translation-based approach outperforms the zero-shot approach by a large margin on the NER task \cite{mayhew-etal-2017-cheap}, such result cannot fully reflect the truth due to the fact that the adopted zero-shot baselines~\cite{tackstrom-etal-2012-cross,bharadwaj-etal-2016-phonologically} are too weak --- they did not introduce multilingual representations from mPTLMs but only used delexicalized features to reduce the language discrepancy.  

In order to faithfully reveal the capabilities of different cross-lingual adaptation approaches, we survey a variety of existing translation-based approaches and systematically compare them with zero-shot approach on three sequence tagging tasks, namely, Named Entity Recognition (\textbf{NER}), Semantic Role Labeling (\textbf{SRL}) and Aspect-Based Sentiment Analysis (\textbf{ABSA}). Note that translation-based approaches for cross-lingual sequence tagging require word-level pseudo labels in the target language, we tailor-make a \textbf{\textit{span-to-span mapping}} component to support assigning pseudo labels for the translated corpus, which aggregates word-based alignment to span-based alignment and propagates labels via aligned spans rather than aligned words (\S~\ref{sec:projection}). Contrary to the previous findings, we show that, even armed with the proposed \textbf{\textit{span-to-span mapping}}, the translation-based approaches are still inferior to the zero-shot approach in most cases. 

Another observation is that performance gains are obtained by performing cross-lingual learning on the combination of the labeled data in the source language and the translated pseudo-labeled data in the target languages for both token-level \cite{fei-etal-2020-cross} and sentence-level tasks \cite{huang-etal-2019-unicoder,conneau-etal-2020-unsupervised}. This observation indicates that even equipping the cross-lingual model with the simplest ``combination'' strategy is beneficial. Presumably, there exists some room for developing more advanced strategies to exploit the multilingual pseudo-parallel data so as to further improve the adaptation performance.
%\citet{huang-etal-2019-unicoder} and \citet{conneau-etal-2020-unsupervised} obtain better cross-lingual SRL results by training models on the union set of the English premise-hypothesis pairs and their translations in 15 target languages. This observation indicates that even equipping the cross-lingual model with the simplest ``combination'' strategy is beneficial. Presumably, there exists some room for developing more advanced strategies to exploit the multilingual pseudo-parallel data so as to further improve the adaptation performance.

Starting from this postulation, we develop Multilingual Warm-Start adaptation (\textsc{\textbf{Mtl-Ws}}), a novel ``\textit{\textbf{warmup-then-adaptation}}'' framework, to optimize the usage of the translated pseudo-labeled data when training the cross-lingual model based on mPTLMs. Concretely, instead of training the cross-lingual model on the entire translated data, we propose to utilize a small number (i.e. warmup steps $\times$ batch size) of pseudo-labeled samples from each translated training set to distill the task-specific knowledge in different target languages.
The obtained knowledge are then aggregated and injected into the cross-lingual model as the multilingual ``warmup'', in order to improve its generalization capability on target languages. With the warm-started cross-lingual model, an existing adaptation approach, e.g., zero-shot approach and target-only translation-based approach, can be further applied to perform the cross-lingual adaptation.

We further evaluate the effectiveness of the proposed \textsc{\textbf{Mtl-Ws}} framework on the aforementioned three sequence tagging tasks, varying the backbone mPTLMs from multilingual BERT~\cite{devlin-etal-2019-bert} to XLM-RoBERTa~\cite{conneau-etal-2020-unsupervised}. The experimental results suggest that our approach surpasses translation-based and zero-shot approaches on almost all language pairs.

\section{Preliminary}
\label{sec:model}
In this section, we first describe the model architecture shared across different adaptation strategies. Then, we present the component supporting the building of pseudo-labeled training set in the target languages, dubbed as ``label projection''\footnote{We use ``label projection'' to avoid confusion with ``annotation projection''~\cite{yarowsky-etal-2001-inducing,das-petrov-2011-unsupervised,akbik-etal-2015-generating}, which generates pseudo-labeled data by utilizing pre-trained model to make automatic annotations and projections on the additional parallel corpus.}.

\subsection{Shared Architecture}
\label{sec:architecture}
\paragraph{Backbone} We regard multilingual pre-trained language models (mPTLMs), usually a deep Transformer~\cite{vaswani2017attention} or deep LSTM architecture~\cite{hochreiter1997long,peters-etal-2018-deep} pre-trained on large-scale multilingual corpus, as the backbone network to calculate multilingual token representations. Thanks to mPTLMs, learning cross-lingual word embeddings~\cite{mikolov2013exploiting,faruqui-dyer-2014-improving,smith2017offline} is no longer prerequisite for cross-lingual transfer.
\paragraph{Task-specific Layer} We adopt a simple feed-forward network with softmax activation, instead of complicated architecture~\cite{akbik-etal-2018-contextual}, as the sequence tagger, following that in~\cite{devlin-etal-2019-bert} and~\cite{conneau-etal-2020-unsupervised}.

Given the input token sequence $\mathrm{\bf x} = \{x_1,\cdots,x_T\}$ of length $T$ (language identifier is ignored for simplicity), we first employ the backbone mPTLM to produce context-aware token representations $\mathrm{\bf H} = \{h_1,\cdots,h_T\} \in \mathbb{R}^{T \times \mathrm{dim}_h}$, where $\mathrm{dim}_h$ denotes the dimension of hidden representations. Then, the token representations are sent to the task layer to perform predictions. Specifically, the probability distribution $p_{\theta}(y_t|x_t) \in \mathbb{R}^{|\mathcal{Y}|}$ (parametrized by $\theta$) over the task-dependent tag set $\mathcal{Y}$ at the $t$-th time step is calculated as follows:
\begin{equation}
    p_{\theta}(y_t|x_t) = \text{softmax} (W h_{t} + b)
\end{equation}
where $W \in \mathbb{R}^{|\mathcal{Y}| \times \mathrm{dim}_h}$ and $b \in \mathbb{R}^{|\mathcal{Y}|}$ are the trainable parameters of the feed-forward network and $\text{softmax}(\cdot)$ refers to the softmax activation function.

\begin{figure*}[t!]
    \centering
    \begin{subfigure}[t]{0.5\textwidth}
        \centering
        \includegraphics[width=0.9\textwidth]{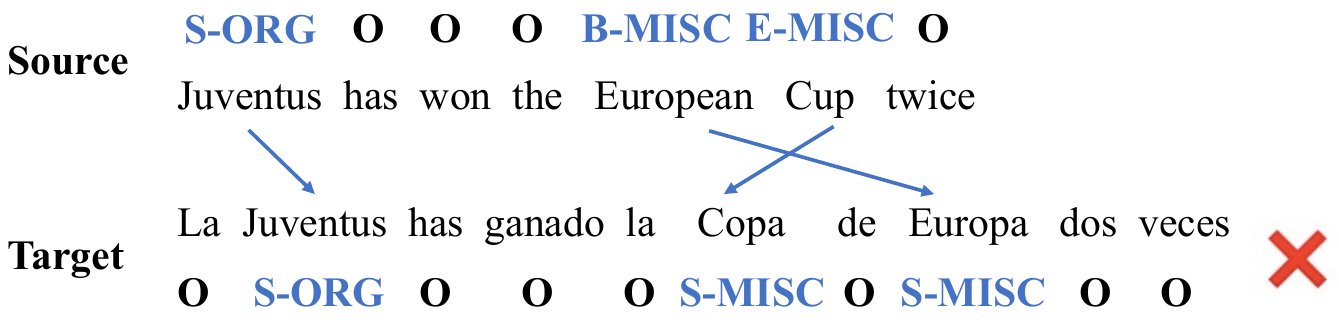}
        \caption{Word-to-word projection}
        \label{fig:word2word}
    \end{subfigure}%
    ~
    \begin{subfigure}[t]{0.5\textwidth}
        \centering
        \includegraphics[width=0.9\textwidth]{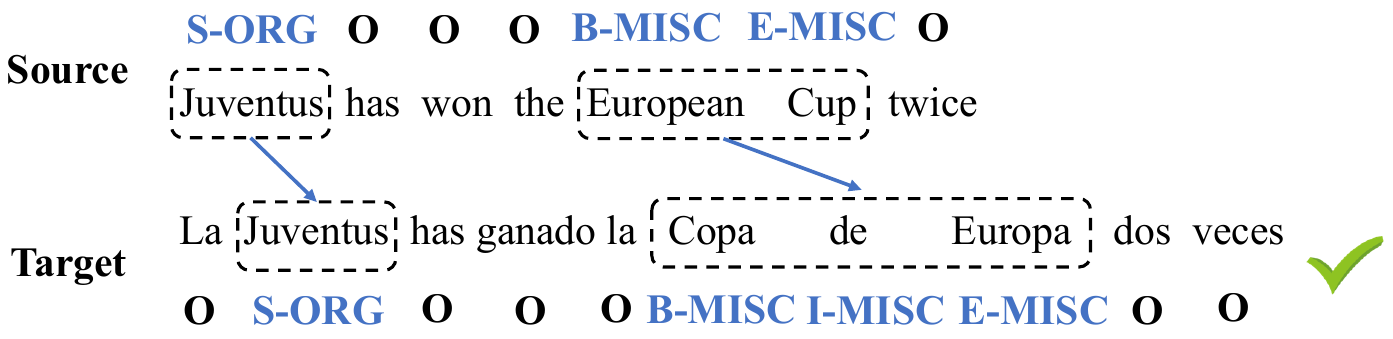}
        \caption{Span-to-span projection}
    \end{subfigure}
    \caption{Example sentence with different label projection strategies. \texttt{BIOES} is adopted as the tagging scheme.}
    \label{fig:label_proj}
\end{figure*}

\subsection{Label Projection for Sequence Tagging}
\label{sec:projection}
Label projection is to propagate the gold standard labels of the source-language training sentences to their translations in the target languages. Since the source sentence is naturally aligned to the translated sentence at sentence-level, label projection for sentence classification or sentence pair classification task is straightforward. When it comes to the sequence tagging task requiring token-level supervision signals, an additional alignment model~\citep{och-ney-2003-systematic,dyer-etal-2013-simple} is needed to produce more fine-grained alignment information. Let $\mathrm{\bf x}^{src} = \{x^{src}_i\}^{T^{src}}_{i=1}$ and $\mathrm{\bf x}^{tgt} = \{x^{tgt}_j\}^{T^{tgt}}_{j=1}$ be the source sentence and the translated sentence in target language respectively. The paired sentences are then sent to the word alignment toolkit to generate word alignment links $\mathrm{\bf a}= \{ a_i \} ^{T^{src}}_{i=1}$, where $a_i \in [1, T^{tgt}] \cup \{\text{NULL}\}$ indicates which target word (or NULL) is the translation of the $i$-th source word. Directly aligning word-level labels~\citep{mayhew-etal-2017-cheap,xie-etal-2018-neural,fei-etal-2020-cross} provides a simple solution for token-level label projection, however, it is fragile to the change of word order and the alignment missing issue, as depicted in Figure~\ref{fig:word2word}. Instead, to improve the alignment quality, we propose ``\textit{\textbf{span-to-span mapping}}'' strategy to transform word-based alignment to span-based alignment and propagate the source labels via the aligned spans. For each gold standard span $\mathrm{\bf x}^{src}_{i:n}$ from the $i$-th source word to the $n$-th source word ($n \geq i$), we locate the aligned span $\mathrm{\bf x}^{tgt}_{j:m}$ in the target language as follows:
\begin{equation}
    \begin{split}
        j &= \min \mathrm{\bf a}_{i:n} \\
        m &= \max \mathrm{\bf a}_{i:n}
    \end{split}
\end{equation}
where $\mathrm{\bf a}_{i:n}$ denotes the alignment links of all source words in $\mathrm{\bf x}^{src}_{i:n}$. Then, we copy the source span label, e.g., \texttt{PER} or \texttt{LOC} in the NER task, to the aligned span and re-generate the boundary label. With this strategy, the post-processing efforts for the potential change of word order are saved and the alignment missing issue is largely alleviated. Note that \textit{\textbf{span-to-span mapping}} is equivalent to word-by-word mapping when 
the source and target spans have the same length and word order.

\section{Zero-shot Adaptation}
\label{sec:zero}
Zero-shot adaptation (\textbf{\textsc{Zero-Shot}}) only fine-tunes model on the source labeled data. The reason that \textbf{\textsc{Zero-Shot}} approach works without explicit guidance in the target language may come from two aspects: 1) the shared tokens across languages serve as the anchors for multilingual generalization~\cite{pires-etal-2019-multilingual}; 2) the deep architecture of mPTLMs can capture some language-independent semantic abstractions~\cite{artetxe2019cross,k2020cross}. The training objective $\mathcal{J}(\theta)$ is as follows:
\begin{equation}
\begin{split}
    \mathcal{J}(\theta) &= \frac{1}{|\mathbb{D}^{src}|}\sum_{(\mathrm{\bf x},\mathrm{\bf y}) \in \mathbb{D}^{src}} \mathcal{L}_{\theta}(\mathrm{\bf x}, \mathrm{\bf y}), \\
    \mathcal{L}_{\theta}(\mathrm{\bf x}, \mathrm{\bf y}) &= -\frac{1}{T}\sum^{T}_{t=1} y^{g}_t \circ \log p_{\theta}(y_t|x_t)
\end{split}
\label{eq:zero_shot}
\end{equation}
where $(\mathrm{\bf x},\mathrm{\bf y}) \in \mathbb{D}^{src}$ denotes a gold standard training sample in the source language and $y^{g}_t \in \mathbb{R}^{|\mathcal{Y}|}$ corresponds to the $t$-th tag (with one-hot encoding) of $\mathrm{\bf y}$. $\circ$ is inner product operation.

\section{Translation-based Adaptation}
\label{sec:translation}
Translation-based adaptation is a group of adaptation approaches exploiting the usage of machine-translated pseudo-labeled data in the target language\footnote{There are two modes of translation-based adaptation, namely, \textbf{Translate-Train} and \textbf{Translate-Test}. The former one guides the learning in the target languages by training on the translated pseudo-labeled data~\cite{banea-etal-2008-multilingual,duh-etal-2011-machine}. While the Translate-Test fine-tunes the cross-lingual model on the source labeled data and then performs inference on the text translated from the target language to the source language~\cite{lambert-2015-aspect,conneau-etal-2018-xnli}. In this paper, we mainly discuss the approaches under \underline{\textbf{Translate-Train}} mode.}. First of all, the source training corpus is translated into the target language. Then, the label projection component armed with the proposed \textit{\textbf{span-to-span mapping}} (see Sec.~\ref{sec:projection}) is employed to propagate labels from the source text to the paired target text along the aligned text segments.

\paragraph{Target-Only}
As the simplest translation-based approach, Target-Only (\textbf{\textsc{To}}) exclusively fine-tunes the model on the translated training set from the specified target language, as done in~\cite{banea-etal-2008-multilingual,duh-etal-2011-machine}. Let $\mathbb{D}^{tgt}$ be the translated corpus, the goal of \textbf{\textsc{To}} is to optimize the following objective:
\begin{equation}
    \mathcal{J}^{\textsc{To}}(\theta) = \frac{1}{|\mathbb{D}^{tgt}|} \sum_{(\mathrm{\bf x}, \hat{\mathrm{\bf y}}) \in \mathbb{D}^{tgt}} \mathcal{L}_{\theta}(\mathrm{\bf x}, \hat{\mathrm{\bf y}}),
    \label{eq:tgt_only}
\end{equation}
where $\hat{\mathrm{\bf y}}$ denotes the pseudo-labels produced by label projection.

\paragraph{Bilingual}
Since machine translation can be regarded as the process of paraphrasing the source text in the target language, Bilingual adaptation approach (\textbf{\textsc{Bilingual}})~\cite{zhang-etal-2019-cross} is developed following the idea of paraphrase-based data augmentation~\cite{wei2018fast,xie2019unsupervised}. Concretely, \textbf{\textsc{Bilingual}} augments the source-language training set with the translated training set in target language. Then, the cross-lingual model is fine-tuned on the augmented bilingual training set. Note that the examples from different languages are not differentiated but treated equally during training. 
The computational process for optimization is as follows:
\begin{equation}
\small
    \mathcal{J}^{\textsc{Bilingual}}(\theta) = \frac{1}{|\mathbb{D}^{src}\cup \mathbb{D}^{tgt}|} \sum_{(\mathrm{\bf x}, \bar{\mathrm{\bf y}}) \in \mathbb{D}^{src}\cup \mathbb{D}^{tgt}} \mathcal{L}_{\theta}(\mathrm{\bf x}, \bar{\mathrm{\bf y}}),
\end{equation}
Here, $\bar{\mathrm{\bf y}}$ is either the ground truth from the source corpus or the pseudo labels from the target corpus.
\paragraph{Multilingual} Introduced by~\citet{yang-etal-2019-paws,huang-etal-2019-unicoder,conneau-etal-2020-unsupervised}, Multilingual adaptation approach (\textbf{\textsc{Mtl}}) is a multilingual extension of \textbf{\textsc{Bilingual}} adaptation, where the model is fine-tuned on the union set of source-language training data and its translations in all of the target languages. Given the task-specific language set $\mathbb{L}$ including the source language and the target languages, the adaptation objective is given below:
\begin{equation}
    \mathcal{J}^{\textsc{Mtl}}(\theta) = \frac{1}{|\bigcup\limits_{\ell \in \mathbb{L}} \mathbb{D}^{\ell}|} \sum_{(\mathrm{\bf x}, \bar{\mathrm{\bf y}}) \in \bigcup\limits_{\ell \in \mathbb{L}} \mathbb{D}^{\ell}} \mathcal{L}_{\theta}(\mathrm{\bf x}, \bar{\mathrm{\bf y}}),
\end{equation}
Similar to \textbf{\textsc{Bilingual}}, the multilingual training examples in \textbf{\textsc{Mtl}} contribute equally to the fine-tuning. One important property of \textbf{\textsc{Mtl}} approach is ``once for all''---performing one-time adaptation for all target languages, greatly reducing the maintenance cost.

\section{Multilingual Warm-Start Adaptation}
The aforementioned \textbf{\textsc{Mtl}} exploits the translated data by training the model on the union set of the data from source language and the machine-translated data in target languages. According to the results in~\citet{fei-etal-2020-cross}, such simple ``combination'' already gives considerable improvement, 
plausibly, designing advanced strategies for incorporating the translated data can further boost the performance.
With this motivation, we propose Multilingual Warm-Start adaptation (\textbf{\textsc{Mtl-Ws}}), a ``\textit{warmup-then-adaptation}'' framework, to optimize the usage of the translated data while inheriting the cross-lingual capability of mPTLMs.

\subsection{Multilingual Warmup}
The first step of our \textbf{\textsc{Mtl-Ws}} is to warm up the model parameters, including the parameters of mPTLMs and those of the task-specific component, with the translated data. Given a subset of translated data in each target language, we perform $Z$ epochs of warmup operation. As observed in~\citet{snyder-etal-2009-adding,ammar-etal-2016-many,ahmad-etal-2019-cross}, the model trained on multiple languages can produce representations that are more suitable for the downstream task or further adaptation, therefore we do warmup on all of the target languages.

Specifically, for the $z$-th epoch, let $\theta_z$ denote the model parameters after warmup, which is calculated from $\theta_{z-1}$ and $|\mathbb{L}|-1$ copies of language specific model parameters $\theta_{z}^{\ell}$:
\begin{equation}
    \theta_z = \theta_{z-1} + \frac{\gamma}{|\mathbb{L}| - 1} \sum_{\ell \in \mathbb{L}\setminus \{src\}} (\theta^{\ell}_{z} - \theta_{z-1}),
\end{equation}
where $\gamma$ denotes the step size of the warmup.
Each $\theta^{\ell}_{z}$ is calculated from $K$ steps of updates by only using the translated data $\mathbb{D}^{\ell}$ of $\ell$:
\begin{equation}
    \theta^{\ell}_{z,k} = \theta^{\ell}_{z,k-1} - \beta \nabla \mathcal{L}_{\theta^{\ell}_{z,k-1}} (\mathrm{\bf x}^{\ell}, \mathrm{\bf \bar{y}}^{\ell}),
\end{equation}
where ${(\mathrm{\bf x}^{\ell}, \mathrm{\bf \bar{y}}^{\ell}) \in \mathbb{D}^{\ell}}$ and ${\theta^{\ell}_{z,0}} \gets {\theta_{z-1}}$.
Since the warmup is performed on the task-specific pseudo-labeled data, the task information and the lexical features in the target languages can be naturally introduced via gradient-based learning on such data.
In \textsc{Mtl-Ws}, we do not directly train the model but separately estimate the task-specific gradients with a small number of gradient update steps on the pseudo-labeled data. The reason why we average the task-specific gradients is that we want to enable the generalizability of the obtained the language-independent task-specific gradients to all of the target languages.

\subsection{Adaptation}
After the warmup encoding of task-specific and language-specific features from the target languages, the approaches mentioned in Sec.~\ref{sec:zero} and~\ref{sec:translation} can be applied to further adapt the warm-started model $\theta_Z$. The multilingual approach, namely \textbf{\textsc{Mtl}}, stands out for its effectiveness and ``once-for-all'' property. However, its training cost is $|\mathbb{L}|-1$ times higher than \textbf{\textsc{Zero-Shot}}, which is computationally prohibitive as the $|\mathbb{L}|$ increases. 
Moreover, our experiments show that even equipped with the span-to-span mapping, these translation-based adaptation approaches are still less effective than \textbf{\textsc{Zero-Shot}} in most cases.
Considering these issues, we employ \textbf{\textsc{Zero-Shot}} approach instead of the multilingual ones to preserve the ``once-for-all'' property while not introducing additional training cost. At the same time, the proposed multilingual warmup mechanism compensates the deficiency that \textbf{\textsc{Zero-Shot}} does not exploit the translated data. 
Similar to Eq.~\ref{eq:zero_shot}, the objective of the adaptation is to minimize the training loss on the source-language data with gold-standard annotations: 
\begin{equation}
    \mathcal{J}^{\textsc{Mtl-Ws}}(\theta_Z) = \frac{1}{|\mathbb{D}^{src}|}\sum_{(\mathrm{\bf x},\mathrm{\bf y}) \in \mathbb{D}^{src}} \mathcal{L}_{\theta_Z}(\mathrm{\bf x}, \mathrm{\bf y}).
\end{equation}

\section{Experiments}
We evaluate the effectiveness of zero-shot approach, translation-based approaches and our Multilingual Warm-Start adaptation approach.

\subsection{Tasks and Datasets}
We conduct experiments on three sequence tagging tasks, namely, Named Entity Recognition (\textbf{NER}), dependency-based Semantic Role Labeling (\textbf{SRL}) and unified Aspect-based Sentiment Analysis (\textbf{ABSA}).
\paragraph{NER} We use datasets from CoNLL-02 and CoNLL-03 NER shared tasks, containing English (en), Spanish (es), German (de) and Dutch (nl)~\cite{tjong-kim-sang-2002-introduction,tjong-kim-sang-de-meulder-2003-introduction} to run NER experiments. 
\paragraph{SRL} We follow the settings of Dependency-based SRL~\cite{roth-lapata-2016-neural,marcheggiani-etal-2017-simple}, whose aim is to identify the syntactic heads of arguments with respect to the given predicate.
As done by~\citet{fei2020cross}, we use datasets from Universal Proposition Bank (UPB)~\cite{akbik-etal-2015-generating}\footnote{\url{https://github.com/System-T/UniversalPropositions}. UPB is built upon Universal Dependency Treebank~\cite{11234/1-1827} and Proposition Bank~\cite{palmer-etal-2005-proposition}.}, containing English (en), French (fr), Spanish (es), German (de), Italian (it), Portuguese (pt), Finnish (fi), to run SRL experiments.
\paragraph{ABSA} We follow the settings of Unified Aspect-based Sentiment Analysis~\cite{mitchell-etal-2013-open,zhang-etal-2015-neural}, which jointly detects the aspect terms mentioned in the reviews and the associated sentiment labels using a single sequence tagging model. We use dataset from SemEval ABSA challenge~\cite{pontiki-etal-2016-semeval}, containing English (en), French (fr), Spanish (es), Turkish (tr), Dutch (nl) and Russian (ru), to run ABSA experiments.

Pre-processing details and dataset statistics are sketched in Appendix.

\subsection{Experimental Settings}
Our cross-lingual models are based on two different mPTLMs, namely, multilingual BERT (\textbf{mBERT}) and XLM-RoBERTa (\textbf{XLM-R}). We use the pre-trained checkpoints\footnote{\texttt{bert-base-multilingual-cased} for \textbf{mBERT} and \texttt{xlm-roberta-base} for \textbf{XLM-R}.} from Huggingface Transformer~\cite{wolf2019transformers} to initialize the backbone mPTLMs. We regard English as the source language and the others as target languages. 

We employ \textit{Google Translate}~\cite{wu2016google,johnson-etal-2017-googles}\footnote{\url{https://translate.google.com/}.} to translate the English training sentences into target languages and \textit{Fast Align}~\cite{dyer-etal-2013-simple}\footnote{\url{https://github.com/clab/fast\_align}.} to generate alignments between the source texts and the translated texts.

\begin{table}[t!]
    \centering
    \resizebox{\columnwidth}{!}{
    \begin{tabular}{l|l|c|c|c|c}
    \Xhline{3\arrayrulewidth}
        \multicolumn{2}{c|}{\textbf{Model}} & \textbf{es} & \textbf{de} & \textbf{nl} & \textbf{Avg.} \\ \hline
        \multirow{5}{*}{\textbf{mBERT}} & \textsc{Zero-Shot} & \textbf{76.14} & 67.92 & 77.10 & 73.72 \\ 
       & \textsc{To} & 69.17 & 63.46 & 70.90 & 67.84 \\ 
       & \textsc{Bilingual} & 71.09 & 67.14 & 71.43 & 69.89 \\ 
       & \textsc{Mtl} & 72.32 & 68.06 & 72.68 & 71.02 \\ 
       %& \textsc{Mtl-Joint} & 70.45 & 65.62 & 71.56 & 69.21 \\ 
       & \textsc{Mtl-Ws} (\textsc{Ours}) & 75.67 & \textbf{71.69} & \textbf{78.31} & \textbf{75.22} \\ \hline \hline
       \multirow{5}{*}{\textbf{XLM-R}} & \textsc{Zero-Shot} & \textbf{77.47} & 68.69 & 78.38 & 74.85 \\ 
       & \textsc{To} & 74.67 & 70.51 & 73.25 & 72.81 \\ 
       & \textsc{Bilingual} & 75.29 & 71.35 & 73.77 & 73.47 \\ 
       & \textsc{Mtl} & 75.38 & 72.10 & 74.60 & 74.03 \\ 
       %& \textsc{Mtl-Joint} & 75.48 & 70.95 & 73.40 & 73.28 \\ 
       & \textsc{Mtl-Ws} (\textsc{Ours}) & 76.73 & \textbf{74.92} & \textbf{78.62} & \textbf{76.76} \\
    \Xhline{3\arrayrulewidth}
    \end{tabular}}
    \caption{Experimental results on NER task (\%). \textbf{Avg.} refers to the averaged $\textsc{F}_1$ score over all of the target languages. We bold the best result in each group.}
    \label{tab:ner_results}
\end{table}

\begin{table}[t!]
    \centering
    \resizebox{\columnwidth}{!}{
    \begin{tabular}{l|l|c|c|c|c|c|c|c}
    \Xhline{3\arrayrulewidth}
        \multicolumn{2}{c|}{\textbf{Model}} & \textbf{fr} &\textbf{es} & \textbf{de} & \textbf{it} & \textbf{pt} & \textbf{fi} & \textbf{Avg.} \\ \hline
        \multirow{5}{*}{\textbf{mBERT}} & \textsc{Zero-Shot} & 42.27 & 45.72 & 48.89 & 51.07 & 47.52 & 41.73 & 46.20 \\ 
       & \textsc{To} & 44.99 & 47.22 & 53.90 & 52.17 & 48.42 & 41.43 & 48.02 \\ 
       & \textsc{Bilingual} & 44.74 & 47.88 & \textbf{54.22} & 53.25 & 49.48 & 41.48 & 48.51 \\ 
       & \textsc{Mtl} & 44.34 & 45.82 & 51.71 & 50.47 & 47.69 & 40.35 & 46.73 \\ 
       %& \textsc{Mtl-Joint} & 41.74 & 43.81 & 50.53 & 48.6 & 45.33 & 40.46 & 45.08 \\ 
       & \textsc{Mtl-Ws} (\textsc{Ours}) & \textbf{45.36} & \textbf{48.21} & 53.20 & \textbf{54.47} & \textbf{50.43} & \textbf{41.95} & \textbf{48.94} \\ \hline \hline
       \multirow{5}{*}{\textbf{XLM-R}} & \textsc{Zero-Shot} & 43.05 & 46.68 & 52.41 & 51.33 & 49.03 & 49.55 & 48.68 \\ 
       & \textsc{To} & 44.99 & 47.55 & 55.20 & 52.25 & 49.54 & 46.41 & 49.32 \\ 
       & \textsc{Bilingual} & 45.06 & 48.75 & \textbf{56.20} & 53.03 & \textbf{51.35} & 46.67 & 50.18 \\ 
       & \textsc{Mtl} & 44.19 & 46.64 & 54.92 & 51.79 & 49.13 & 44.18 & 48.48 \\ 
       %& \textsc{Mtl-Joint} & 44.08 & 45.41 & 53.94 & 49.34 & 47.77 & 45.10 & 47.61 \\ 
       & \textsc{Mtl-Ws} (\textsc{Ours}) & \textbf{45.74} & \textbf{49.76} & 54.50 & \textbf{53.80} & 50.59 & \textbf{50.73} & \textbf{50.85} \\
    \Xhline{3\arrayrulewidth}
    \end{tabular}}
    \caption{Experimental results on SRL task (\%). We bold the best result in each group.}
    \label{tab:srl_results}
\end{table}

\begin{table}[t!]
    \centering
    \resizebox{\columnwidth}{!}{
    \begin{tabular}{l|l|c|c|c|c|c|c}
    \Xhline{3\arrayrulewidth}
        \multicolumn{2}{c|}{\textbf{Model}} & \textbf{fr} &\textbf{es} & \textbf{tr} & \textbf{nl} & \textbf{ru} & \textbf{Avg.} \\ \hline
        \multirow{5}{*}{\textbf{mBERT}} & \textsc{Zero-Shot} & 45.60 & 57.32 & 26.58 & 42.68 & 36.01 & 41.64 \\ 
       & \textsc{To} & 40.76 & 50.74 & 22.04 & 47.13 & 41.67 & 40.47 \\ 
       & \textsc{Bilingual} & 41.00 & 51.23 & 22.64 & 49.72 & 43.67 & 41.65 \\ 
       & \textsc{Mtl} & 40.72 & 54.14 & 25.77 & 49.06 & 43.89 & 42.72 \\ 
       %& \textsc{Mtl-Joint} & 40.44 & 51.16 & 24.45 & 48.57 & 43.68 & 41.66 \\ 
       & \textsc{Mtl-Ws} (\textsc{Ours}) & \textbf{46.93} & \textbf{58.18} & \textbf{29.78} & \textbf{49.87} & \textbf{44.88} & \textbf{45.93} \\ \hline \hline
       \multirow{5}{*}{\textbf{XLM-R}} & \textsc{Zero-Shot} & 56.43 & 67.10 & \textbf{46.21} & 59.03 & 56.80 & 57.11 \\ 
       & \textsc{To} & 47.00 & 58.10 & 40.24 & 56.19 & 50.34 & 50.37 \\ 
       & \textsc{Bilingul} & 49.34 & 61.87 & 41.44 & 58.64 & 52.89 & 52.84 \\ 
       & \textsc{Mtl} & 52.80 & 63.56 & 43.04 & 60.37 & 55.67 & 55.09 \\ 
       %& \textsc{Mtl-Joint} & 48.57 & 60.63 & 43.51 & 58.70 & 53.19 & 52.92 \\ 
       & \textsc{Mtl-Ws} (\textsc{Ours}) & \textbf{57.96} & \textbf{68.60} & 45.58 & \textbf{61.24} & \textbf{59.74} & \textbf{58.62} \\
    \Xhline{3\arrayrulewidth}
    \end{tabular}}
    \caption{Experimental results on ABSA task (\%). We bold the best result in each group.} 
    \label{tab:absa_results}
\end{table}

As with tagging schemes, we adopt \texttt{BIOES} for NER, \texttt{OI} for SRL and \texttt{BIOES} for ABSA\footnote{Please refer to~\cite{tjong-etal-1999-representing} for more details about these tagging schemes.}, as done in \citep{lample-etal-2016-neural}, \citep{marcheggiani-etal-2017-simple} and \citep{li-etal-2019-exploiting} respectively.  All of these tasks follow span-based evaluation\footnote{Output is correct if and only if it exactly matches the span boundaries and span label.%\lx{MARK}
} and micro-$\mathrm{F}_1$ score is reported. The numbers in Table~\ref{tab:ner_results}-\ref{tab:absa_results} are the averaged results over 5 runs with different random seeds, as suggested in~\citep{reimers-gurevych-2017-reporting}. We employ Adam~\cite{kingma2014adam} for model optimization.

Since our aim is not to build a new state-of-the-art model but investigate effective strategies for exploiting the translated pseudo-labeled data, thus, we do not compare with the previous best models~\cite{wu2019enhanced,wang2020crosslingual,fei2020cross,saiful2020multimix} on cross-lingual NER or cross-lingual SRL tasks. In principle, our \textbf{\textsc{Mtl-Ws}}, a model-agnostic framework, can be packed with such mPTLM-based models to further improve the adaptation performance, which is a worthwhile direction to explore. 

Concrete details with respect to the experiment are presented in Appendix. Note that, for fair comparison, we keep the parameter settings and the model selection strategies of the proposed \textbf{\textsc{Mtl-Ws}} identical to those of the compared adaptation approaches.

\subsection{Result Discussions}
Table~\ref{tab:ner_results}, Table~\ref{tab:srl_results} and Table~\ref{tab:absa_results} present the experimental results for NER, SRL and ABSA respectively.

\subsubsection{Main Results}
\label{sec:main_results}
As shown in Tables~\ref{tab:ner_results}-\ref{tab:absa_results}, the proposed \textbf{\textsc{Mtl-Ws}} approach achieves the best averaged $\textsc{F}_1$ scores on different multilingual pre-trained language models (mPTLMs), demonstrating its effectiveness on cross-lingual sequence tagging. Comparing with the best translation-based approaches, namely, \textbf{\textsc{Mtl}}, our \textbf{\textsc{Mtl-Ws}}, which
injects the language- \& task-specific knowledge into the cross-lingual model via the warmup mechanism, makes better use of the multilingual training set 
and brings in 4.2\%, 2.2\% and 3.2\% absolute gains for the mBERT-based NER model, SRL model and ABSA model, respectively. 

\textbf{\textsc{Mtl-Ws}} can be seen as applying \textbf{\textsc{Zero-Shot}} approach on the warm-started model parameters. By exploiting the source labeled data as well as the multilingual pseudo-labeled data, the cross-lingual model is better adapted to the target languages, especially on the ABSA task, the tailor-made multilingual warmup boosts the averaged score by $\sim$10\% with mBERT as backbone.

We also observe that changing the backbone from mBERT to XLM-R drastically advances the $\textsc{F}_1$ score of \textbf{\textsc{Mtl-Ws}} on ABSA (45.9$\Rightarrow$58.6). It is presumably because XLM-R pre-trained on much larger corpus is capable of capturing rich lexical variation and handling potential spelling errors from the informal texts. Meanwhile, the performance gain on NER (75.2$\Rightarrow$76.8), where the data is built upon the more formal Reuters news corpus, is not that significant, suggesting that mBERT is still a promising choice when handling the formal textual input such as news articles. 

\subsubsection{Zero-shot or Using Translation?}
According to the results on the NER and ABSA tasks, \textbf{\textsc{Zero-Shot}} without any guidance in the target language outperforms \textbf{\textsc{To}} on the majority of language pairs. Compared to \textbf{\textsc{Mtl}}, \textbf{\textsc{Zero-Shot}} obtains comparable or even better averaged $\textsc{F}_1$ score. Such findings are contrary to those from cross-lingual sequence (or sequence pair) classification tasks. We attribute this to the lower tolerance of NER model and ABSA model to the translation errors. More specifically, the issues of under-translation and mis-translation have the tendency to lose the expected translations of the aspect/entity mention, which will hinder the label projection for these two tasks requiring the token-level labels. While the yielding side-effect on the sequence-level tasks is less severe. Besides, the translation-based approaches on sequence tagging rely on unsupervised aligners to produce alignment links, which may introduce additional noises, even using our proposed span-to-span projection.

It is notable that the translation-based \textbf{\textsc{To}} and \textbf{\textsc{Mtl}} are still superior to \textbf{\textsc{Zero-Shot}} on the SRL task. The probable reason is that the lexical transfer in \textbf{\textsc{Zero-Shot}} is insufficient for discovering the internal structure of sentence. At the same time, the translation-based counterparts can implicitly generate the predicate-argument dependencies 
in the translated text via the projection of semantic role labels, which directly helps the learning of semantic roles in the target language.

\begin{table}[]
    \centering
    \resizebox{\columnwidth}{!}{
    \begin{tabular}{l|ccc||cc}
    \Xhline{3\arrayrulewidth}
        & \multicolumn{3}{c||}{\textit{English}} & \multicolumn{2}{c}{\textit{Low-Resource Languages}} \\
         & NER & SRL & ABSA & SRL (fi) & ABSA (tr)  \\ \hline
        mBERT & 91.35 & 82.37 & 66.67 & 71.77 & 43.66  \\
        XLM-R & 91.46 & 84.03 & 73.74 & 76.7 & 65.73 \\ \hline \hline
        $\Delta$ & 0.09 & 1.66 & 7.07 & 4.93 & 22.07 \\ 
    \Xhline{3\arrayrulewidth}
    \end{tabular}}
    \caption{Monolingual performances (F$_1$).}
    \label{tab:monolingual}
\end{table}

\subsubsection{Further Discussions on mPTLMs}
As discussed in Sec. \ref{sec:main_results}, the performance gap between mBERT and XLM-R is exceptionally large on the cross-lingual ABSA task. Here, we attempt to empirically track the sources that may contribute to this situation. We first conduct monolingual experiments on English with results shown in Table~\ref{tab:monolingual}. As can be seen, the $\textsc{F}_1$ scores between mBERT and XLM-R are very close when applied to the formal NER/SRL texts, while on the ABSA task, the former is far behind the latter, suggesting that handling informal texts with rich morphology changes is challenging for mBERT even in the most resource-rich language, i.e. EN. We also evaluate the monolingual performances of mPTLMs on two low-resource languages: Turkish (tr) and Finnish (fi)\footnote{According to the statistics in~\url{https://meta.wikimedia.org/wiki/List_of_Wikipedias}, Finnish ranked the 25th and Turkish ranked the 32nd.} and the right side of Table~\ref{tab:monolingual} depicts the evaluation results. Regarding the monolingual results on the SRL task where the mPTLMs do not suffer from the misspellings too much, the gap between mBERT and XLM-R on $\textsc{F}_1$ score is still enlarged ($\Delta$: 1.66$\Rightarrow$4.93) when varying English to Finnish, indicating that the mBERT cannot provide satisfactory representations in low-resource language for the predictions. In summary, the fragility to informal texts and the limited modeling power in low-resource language are two factors hindering the cross-lingual transfer of mBERT on sequence tagging tasks.

\begin{figure}[!t]
    \centering
    \includegraphics[width=.4\textwidth]{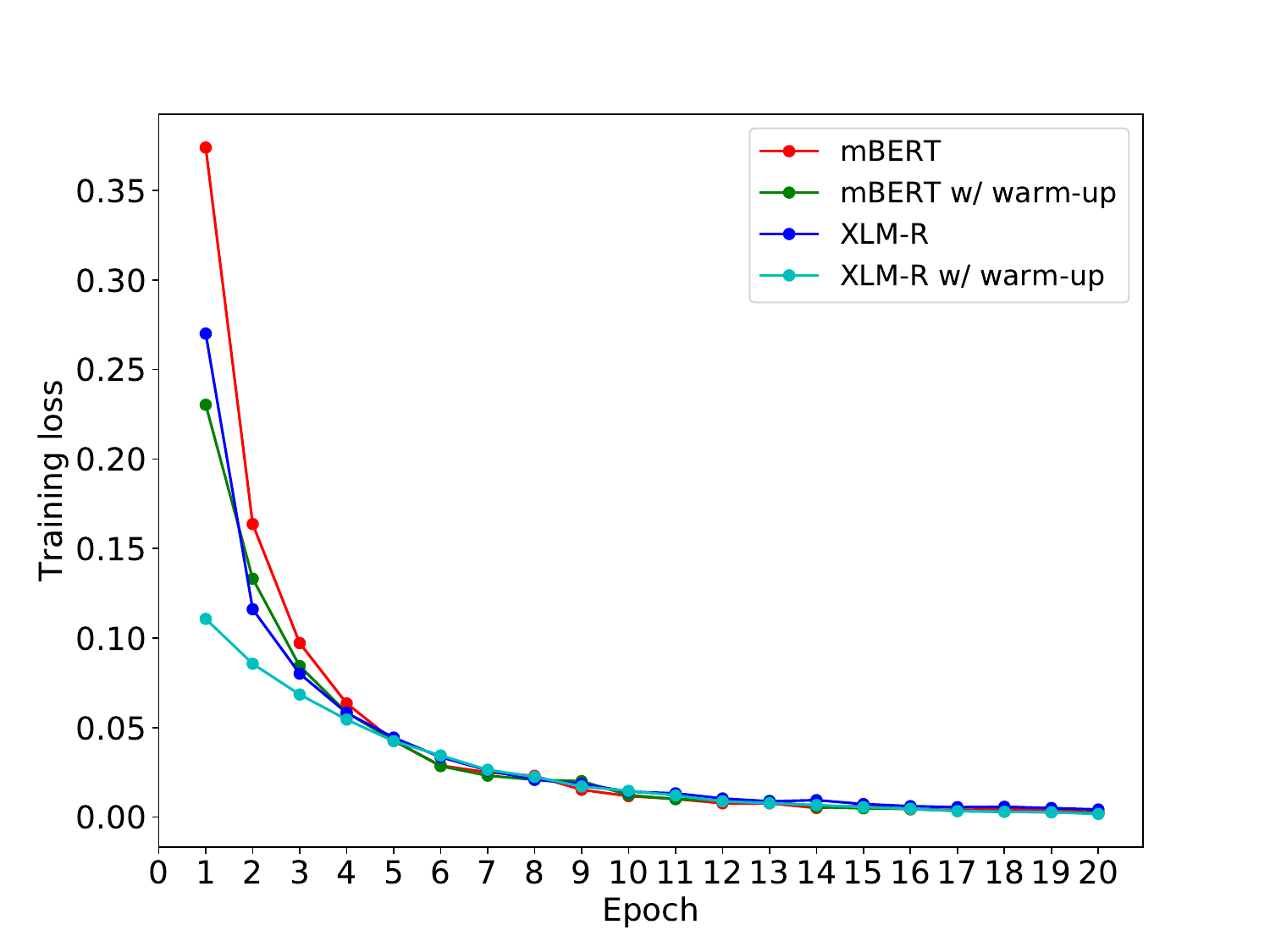}
    \caption{Effects of warmup mechanism on convergence.
    }
    \label{fig:warmup}
\end{figure}
%\vspace{-1mm}

\subsection{Effects of Warmup on Convergence}
We further investigate the effects of the proposed warmup mechanism on the cross-lingual adaptation. As shown in Fig.~\ref{fig:warmup}, after introducing the warmup mechanism, both mBERT and XLM-R converge faster at the first few epochs.
Such property is especially useful when the training data is large and the computational power is limited.

\subsection{Effects of \textit{Span-to-Span Mapping}}
As mentioned in Sec.~\ref{sec:projection}, the proposed \textbf{\textit{span-to-span}} mapping is more robust than the label projection via aligned words. Here, we assess this property on NER. We fix the target language as Spanish and compare the downstream adaptation performances of \textbf{\textsc{To}} with different label projection strategies. As shown in Table~\ref{tab:label_proj}, replacing our span-to-span strategy with the word-to-word strategy leads to remarkable decline of $\textsc{F}_1$ scores, suggesting that the quality of the pseudo-labeled data is unacceptable without considering the potential change of word order and missing alignment. The experimental results also demonstrate that our \textbf{\textit{span-to-span}} mapping, a simple heuristic strategy, can alleviate the problem to some extent.

\begin{table}[]
    \centering
    \resizebox{0.9\columnwidth}{!}{
    \begin{tabular}{l|c|c|c}
     \Xhline{3\arrayrulewidth}
     Backbone & span-to-span & word-to-word  & $\Delta_{\text{LP}}$ \\ \Xhline{3\arrayrulewidth}
     \textbf{mBERT} & 69.17 & 58.46 & -10.71 \\  \Xhline{3\arrayrulewidth}
     \textbf{XLM-R} & 74.67 & 61.70 & -12.97  \\  
     \Xhline{3\arrayrulewidth}
    \end{tabular}}
    \caption{Performances of \textbf{\textsc{To}} on NER with different label projection strategies.}
    \label{tab:label_proj}
\end{table}
\vspace{-2mm}

\section{Related Work}

\paragraph{Unsupervised Multilingual Pre-training} With the help of deep neural architectures~\cite{vaswani2017attention,peters-etal-2018-deep} and unsupervised representation learning techniques, multilingual Pre-trained Language Models (mPTLMs)~\cite{che-etal-2018-towards,devlin-etal-2019-bert,lample2019cross,mulcaire-etal-2019-polyglot,conneau-etal-2020-unsupervised} greatly advances the state-of-the-art for a series of zero-shot cross-lingual natural language understanding tasks~\cite{prettenhofer-stein-2010-cross,schwenk-li-2018-corpus,conneau-etal-2018-xnli,zeman-etal-2018-conll,liu-etal-2019-xqa}. Exploiting mPTLMs for cross-lingual learning usually involves two phases: general-purpose pre-training on multilingual corpus and task-specific adaptation. Since the cost of improving the former is not affordable to the majority of the NLP community, the researchers turn to explore more effective adaptation approaches for better performances.        

\paragraph{Cross-lingual Adaptation} The existing studies of mPTLM-based cross-lingual adaptation can be categorized into two lines: 1) zero-shot approach; and 2) translation-based approach. The zero-shot approach~\cite{pires-etal-2019-multilingual,wu-dredze-2019-beto,keung-etal-2019-adversarial} solely fine-tunes the cross-lingual model on the source labeled data and there is no explicit lexical transfer except the shared tokens across different languages.
%\isaac{captured during the pre-training of the mPTM}.
Instead, the translation-based approach borrows machine translation system to perform explicit transfer and tailor-makes some strategies to make use of the translated training set in target languages. Some ideas such as paraphrase-based data
augmentation~\cite{wei2018fast,xie2019unsupervised} and multi-task learning~\cite{caruana1997multitask,ruder2017overview} have been adopted in translation-based approaches and achieve encouraging results on cross-lingual adaptation~\cite{artetxe2019massively,yang-etal-2019-paws,huang-etal-2019-unicoder,conneau-etal-2020-unsupervised,cao2020multilingual}. The Multilingual Warm-Start adaptation (\textbf{\textsc{Mtl-Ws}}) proposed in this paper can be viewed as a hybrid of the zero-shot approach and the translation-based approach.

\section{Conclusions}
In this paper, we re-examine the effectiveness of the existing cross-lingual adaptation approaches for multilingual pre-trained language model (mPTLMs) on sequence tagging tasks. We show how to perform label projection for enabling the translation-based approaches in sequence tagging task, and develop a \textbf{\textit{span-to-span}} mapping strategy to handle the potential change of word order and alignment missing issue. Moreover, we propose Multilingual Warm-Start (\textsc{Mtl-Ws}) adaptation to further enhance the cross-lingual transfer. Specifically, we tailor-make a warmup mechanism to distill the knowledge from the translated training sets and inject the obtained knowledge into the cross-lingual model. As the multilingual warmup is able to compensate the deficiency of not exploiting the translated data, \textbf{Zero-Shot} approach is then applied to preserve the ``\textit{onece for all}'' property while not increasing the training cost. Through the comprehensive experiments over nine languages from three sequence tagging tasks, we demonstrate that the proposed \textsc{Mtl-Ws} works well across different languages without any task-specific design for the downstream model. We also demonstrate that the superiority of our approach to other adaptation approaches is generally consistent across different mPTLMs.  

\bibliography{emnlp2021}
\bibliographystyle{acl_natbib}

\appendix

\begin{table*}[!t]
    \centering
    {%
    \begin{tabular}{l|ccc|ccc}
    \Xhline{3\arrayrulewidth}
    \multirow{2}{*}{\textbf{Hyper-params}} & \multicolumn{3}{c}{\textbf{mBERT}} & \multicolumn{3}{|c}{\textbf{XLM-R}} \\ \cline{2-7}
    & NER & SRL & ABSA & NER & SRL & ABSA \\ \hline
    $Z$ & 100 & 50 & 100 & 50 & 100 & 100 \\ %\hline
    $K$ & 5 & 30 & 20 & 20 & 30 & 20 \\ %\hline
    $\gamma$ & \multicolumn{3}{c}{5e-5} & \multicolumn{3}{|c}{5e-5} \\ %\hline
    $\beta$ & \multicolumn{3}{c}{1e-3} & \multicolumn{3}{|c}{1e-3}  \\ %\hline
    learning rate & \multicolumn{3}{c}{5e-5} & \multicolumn{3}{|c}{1e-5} \\ %\hline
    batch size & \multicolumn{3}{c}{8} & \multicolumn{3}{|c}{4}  \\ %\hline
    \Xhline{3\arrayrulewidth}
    \end{tabular}}
    \caption{The general experimental settings. \textbf{mBERT}/\textbf{XLM-R} here refers to the cross-lingual model with a mBERT/XLM-R backbone.}
    \label{tab:settings}
\end{table*}

\section{Settings}
\subsection{General Settings}
For each task, we only search the best number of warmup iterations $Z$, the best number of gradient updates $k$ within each warmup iteration and the best learning rate for the adaptation on the English development set. The hyper-parameter ranges are as follows: the number of warmup iterations $Z$ \{50, 75, 100\}; the number of warmup steps $k$ \{5, 10, 20, 30\}; learning rate \{5e-4, 1e-5, 2e-5, 5e-5\}. Batch size for mBERT-based model is set as 8 and that for XLM-R-based model is set as 4, due to the limit of GPU memory. Other hyper-parameters are empirically selected. The step size for warmup, namely $\gamma$, and that for gradient-based training on the target language, namely $\beta$, are set as 5e-5 and 1e-3 respectively. The finalized parameter settings are given in Table~\ref{tab:settings}.

\begin{table*}[]
    \centering
    \resizebox{0.8\textwidth}{!}{
    \begin{tabular}{l|cc|cc|cc}
    \Xhline{3\arrayrulewidth}
    \multirow{2}{*}{Language} & \multicolumn{2}{c|}{Train} & \multicolumn{2}{c|}{Dev} & \multicolumn{2}{c}{Test}   \\ \cline{2-7}
        & \# \texttt{sent.} & \# \texttt{entity} & \# \texttt{sent.} & \# \texttt{entity} & \# \texttt{sent.} & \# \texttt{entity} \\ \hline
        en & 14041 & 23499 & 3250 & 5942 & 3453 & 5648 \\ 
        es & 8323 & 18798 & 1915 & 4351 & 1517 & 3558 \\
        de & 12152 & 11851 & 2867 & 4833 & 3005 & 3673 \\ 
        nl & 15806 & 13344 & 2895 & 2616 & 5195 & 3941 \\
    \Xhline{3\arrayrulewidth}
    \end{tabular}}
    \caption{Statistics of NER datasets.}
    \label{tab:ner_stat}
\end{table*}

\begin{table}[]
    \centering
    \resizebox{1\columnwidth}{!}{
    \begin{tabular}{l|cc|cc}
    \Xhline{3\arrayrulewidth}
    \multirow{2}{*}{Language} & \multicolumn{2}{c|}{Train} & \multicolumn{2}{c}{Test}   \\ \cline{2-5}
        & \# \texttt{sent.} & \# \texttt{aspect} & \# \texttt{sent.} & \# \texttt{aspect} \\ \hline
        en & 2000 & 2507 & 676 & 859  \\ 
        fr & 1733 & 2530 & 696 & 950  \\
        es & 2070 & 2720 & 881 & 1072  \\ 
        tr & 1104 & 1535 & 144 & 159 \\
        nl & 1711 & 1860 & 575 & 613 \\
        ru & 3490 & 4022 & 1209 & 1300 \\
    \Xhline{3\arrayrulewidth}
    \end{tabular}}
    \caption{Statistics of ABSA datasets.}
    \label{tab:absa_stat}
\end{table}

\subsection{Task-specific Settings}
\label{appendix:example}
\paragraph{NER} Table~\ref{tab:ner_stat} depicts the statistics for NER datasets. For the NER experiments, we truncate the maximum length of sub-word sequence to 128 and adopt the gold standard train/dev/test split. We train mBERT-based NER model up to 3000 steps and conduct model selection after 2000 steps, where we evaluate the performance on development set every 200 steps. For those based on XLM-R, We train them up to 6,000 steps. Similarly, after training 4000 steps, we test the model performance on development set every 400 steps to select the model for prediction.

\begin{table*}[]
    \centering
    \resizebox{0.8\textwidth}{!}{
    \begin{tabular}{l|cc|cc|cc}
    \Xhline{3\arrayrulewidth}
    \multirow{2}{*}{Language} & \multicolumn{2}{c|}{Train} & \multicolumn{2}{c|}{Dev} & \multicolumn{2}{c}{Test}   \\ \cline{2-7}
        & \# \texttt{sent.} & \# \texttt{pred.} & \# \texttt{sent.} & \# \texttt{pred.} & \# \texttt{sent.} & \# \texttt{pred.} \\ \hline
        en & 10908 & 40149 & 1632 & 4892 & 1634 & 4700 \\
        fr & 14554 & 29263 & 1596 & 3043 & 298 & 635 \\
        es & 28492 & 73254 & 3206 & 8274 & 1995 & 5437 \\
        de & 14418 & 21261 & 799 & 1180 & 977 & 1317 \\ 
        it & 12837 & 25621 & 489 & 1009 & 489 & 1014 \\
        pt & 7494 & 16842 & 938 & 2071 & 936 & 2107 \\
        fi & 12217 & 25584 & 716 & 1477 & 648 & 1451 \\
    \Xhline{3\arrayrulewidth}
    \end{tabular}}
    \caption{Statistics of SRL datasets. \# \texttt{pred.} refers to the number of predicates.}
    \label{tab:srl_stat}
\end{table*}

\paragraph{SRL} Table~\ref{tab:srl_stat} depicts the statistics for SRL datasets. Apart from the text, we also feed the Part-of-speech (POS) tags provided by UPB as the input of the model. We use stanza~\cite{qi2020stanza}\footnote{\url{https://github.com/stanfordnlp/stanza.}} to produce POS tags for the translated data. For the model based on mBERT, we train them up to 10000 steps and conduct model selection every 1000 steps. For the models based on XLM-R, we train them up to 20000 steps and conduct model selection every 2000 steps.

\paragraph{ABSA} The dataset statistics are shown in Table~\ref{tab:absa_stat}. Since there is no official development set, we randomly sample 20\% training samples and regard them as development data. We do not perform truncation for the sentences in ABSA datasets. Following~\citet{li-etal-2019-exploiting}, we train the model up to 1,500 steps and conduct model selection after 1000 steps. For model selection, we evaluate the performance on development set every 100 steps.

\end{document}